\documentclass[twocolumn,10pt]{asme2ej}

\usepackage{graphicx}
\usepackage{amsmath}
\usepackage{amssymb}
\usepackage{booktabs}
\usepackage{xcolor}
\usepackage{algorithm}
\usepackage{algpseudocode}
\definecolor{mylavender}{RGB}{160,130,210}
\newcommand{\carrie}[1]

\usepackage[pagebackref,breaklinks,colorlinks]{hyperref}

\usepackage[capitalize]{cleveref}
\crefname{section}{Sec.}{Secs.}
\Crefname{section}{Section}{Sections}
\Crefname{table}{Table}{Tables}
\crefname{table}{Tab.}{Tabs.}

\begin{document}
\title{Parametric-ControlNet: Multimodal Control in Foundation Models for Precise Engineering Design Synthesis}

\author{Rui Zhou
    \affiliation{
	Department of Mechanical Engineering\\
	Massachusetts Institute of Technology\\
	Cambridge, Massachusetts, 02139\\
    Email: zhourui@mit.edu
    }	
}

\author{Yanxia Zhang
    \affiliation{
	Toyota Research Institute\\
	Los Altos, CA 94022\\
    Email: yanxia.zhang@tri.global
    }	
}
\author{Chenyang Yuan
    \affiliation{
	Toyota Research Institute\\
	Los Altos, CA 94022\\
    Email: chenyang.yuan@tri.global
    }	
}
\author{Frank Permenter
    \affiliation{
	Toyota Research Institute\\
	Los Altos, CA 94022\\
    Email: frank.permenter@tri.global
    }	
}
\author{Nikos Arechiga
    \affiliation{
	Toyota Research Institute\\
	Los Altos, CA 94022\\
    Email: nikos.arechiga@gmail.com
    }	
}
\author{Matt Klenk
    \affiliation{
	Toyota Research Institute\\
	Los Altos, CA 94022\\
    Email: matt.klenk@tri.global
    }	
}

\author{Faez Ahmed
     \affiliation{
	Department of Mechanical Engineering\\
	Massachusetts Institute of Technology\\
	Cambridge, Massachusetts, 02139\\
    Email: faez@mit.edu
    }	
}

\maketitle
\begin{abstract}
   This paper introduces a generative model designed for multimodal control over text-to-image foundation generative AI models such as Stable Diffusion, specifically tailored for engineering design synthesis. Our model proposes parametric, image, and text control modalities to enhance design precision and diversity. Firstly, it handles both partial and complete parametric inputs using a diffusion model that acts as a design autocomplete co-pilot, coupled with a parametric encoder to process the information. Secondly, the model utilizes assembly graphs to systematically assemble input component images, which are then processed through a component encoder to capture essential visual data. Thirdly, textual descriptions are integrated via CLIP encoding, ensuring a comprehensive interpretation of design intent. These diverse inputs are synthesized through a multimodal fusion technique, creating a joint embedding that acts as the input to a module inspired by ControlNet. This integration allows the model to apply robust multimodal control to foundation models, facilitating the generation of complex and precise engineering designs. This approach broadens the capabilities of AI-driven design tools and demonstrates significant advancements in precise control based on diverse data modalities for enhanced design generation.
\end{abstract}
\section{Introduction}
\begin{figure*}[t]
\label{fig:stable_diffusion_fail1}
\centering
\includegraphics[width=0.9\textwidth]{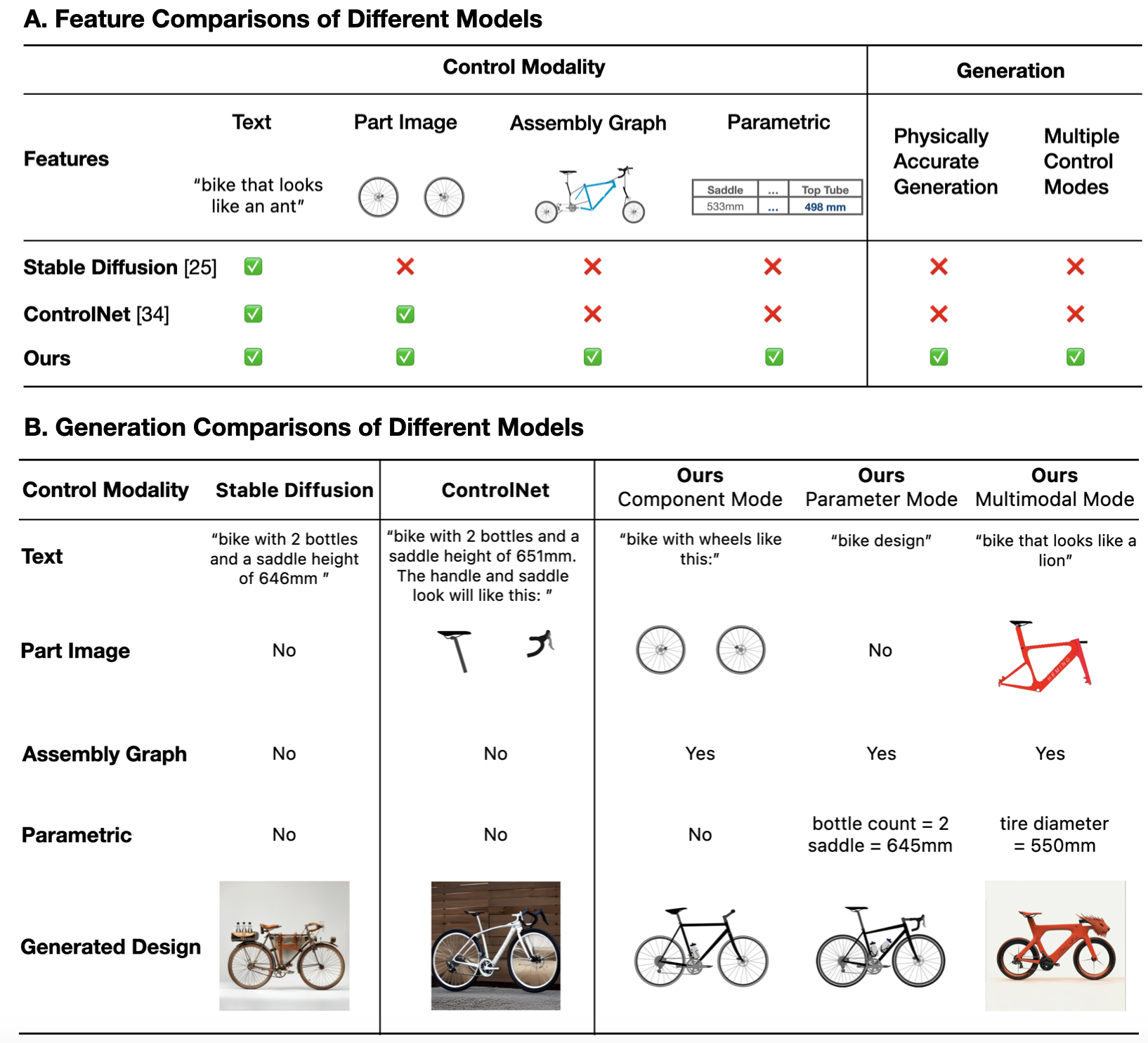}
\caption{Comparisons of features and generation results for different models. In Subfigure A, we show the additional control modalities and features that our model offers in comparison to the state-of-the-art models. in Subfigure B, we compare the performance in precise multimodal control and generation quality for different models.}
\end{figure*}
Engineering design is a complex process that involves the creation, analysis, and optimization of products, systems, and processes to meet specific requirements. Traditionally, engineering design has relied on human expertise, creativity, and problem-solving skills to navigate the vast design space and arrive at optimal solutions. However, with the rapid advancement of artificial intelligence (AI) techniques, there is a growing opportunity toenhance the engineering design process through the use of AI-driven tools and methodologies.
Generative models, in particular, have emerged as a promising avenue for bridging the gap between AI and engineering design. These models, which learn to generate new data points based on patterns and structures observed in training data, have the potential to unlock new possibilities in design exploration, optimization, and automation. By leveraging the power of deep learning and probabilistic modeling, generative models can help designers explore a wider range of design alternatives, uncover novel design concepts, and streamline the design process. However, despite the success of generative models in domains such as art, music, and natural image generation~\cite{dhariwal2021diffusion}, their applications in engineering design have been limited. This is due to several factors. These limitations are caused by three significant factors: (a) designers are unable to precisely control the generated content, (b) generative models have difficulties in understanding performance metrics and physical properties, and (c) generative models often struggle with complex engineering designs~\cite{review}.  For example, models like Stable Diffusion~\cite{sd} excel at generating realistic images based on textual descriptions but struggle to generate designs that accurately reflect parametric inputs or assembly constraints as shown in figure~\ref{fig:stable_diffusion_fail1}. For example, all four examples by stable diffusion fail to conform to the requirement of 2 water bottles.
In contrast, our model successfully generates diverse designs that conform to the requirement. This exemplifies the problems with existing text-to-image (T2I) foundation models. Specifically, they can generate novel and realistic-looking images but fail to generate feasible and functional designs that follow engineering requirements such as parametric and assembly constraints.

To address these limitations and fully realize the potential of generative models in engineering design, we introduce a generative model architecture that enables precise multimodal control over foundation T2I models. In addition to text input, our model incorporates three distinct control modalities tailored for engineering design: parametric inputs, assembly graphs, and component inspiration images. Through multimodal control over foundation models, our approach enables designers to generate designers conditioned new types of information.
The contributions of this paper are threefold:
\begin{enumerate}
    \item \textbf{Multimodal Control Architecture:} We propose an architecture that enables control over foundation models using the following modalities: parametric data, assembly graphs, component images, and text. We accomplish this by adding a ControlNet-like module with the following components: a diffusion-based parametric autocompletion and encoding module, a component encoder, CLIP encoder, and a multimodal fusion module. This architecture provides designers with enhanced control over generated designs, ensuring alignment with specifications, constraints, and preferences.
    \item \textbf{Performance Evaluation:} We demonstrate the unique capabilities of our model in engineering design generation through extensive experiments and evaluations. Unlike state-of-the-art models, our model can generate designs that closely adhere to provided parametric specifications, assembly constraints, and creative prompts while maintaining high visual quality and diversity. We showcase the model's ability to handle complex engineering design tasks that are beyond the scope of existing generative models, such as generating designs with specific functional requirements and spatial constraints.
    \item \textbf{Applications in Engineering Design:} Our model provides new possibilities for engineering design generation and exploration by leveraging multimodal control. We propose a pipeline that designers can use for design exploration. 
\end{enumerate}
\section{Related Works}
\begin{figure*}[t]
    \centering
    \includegraphics[width=14cm]{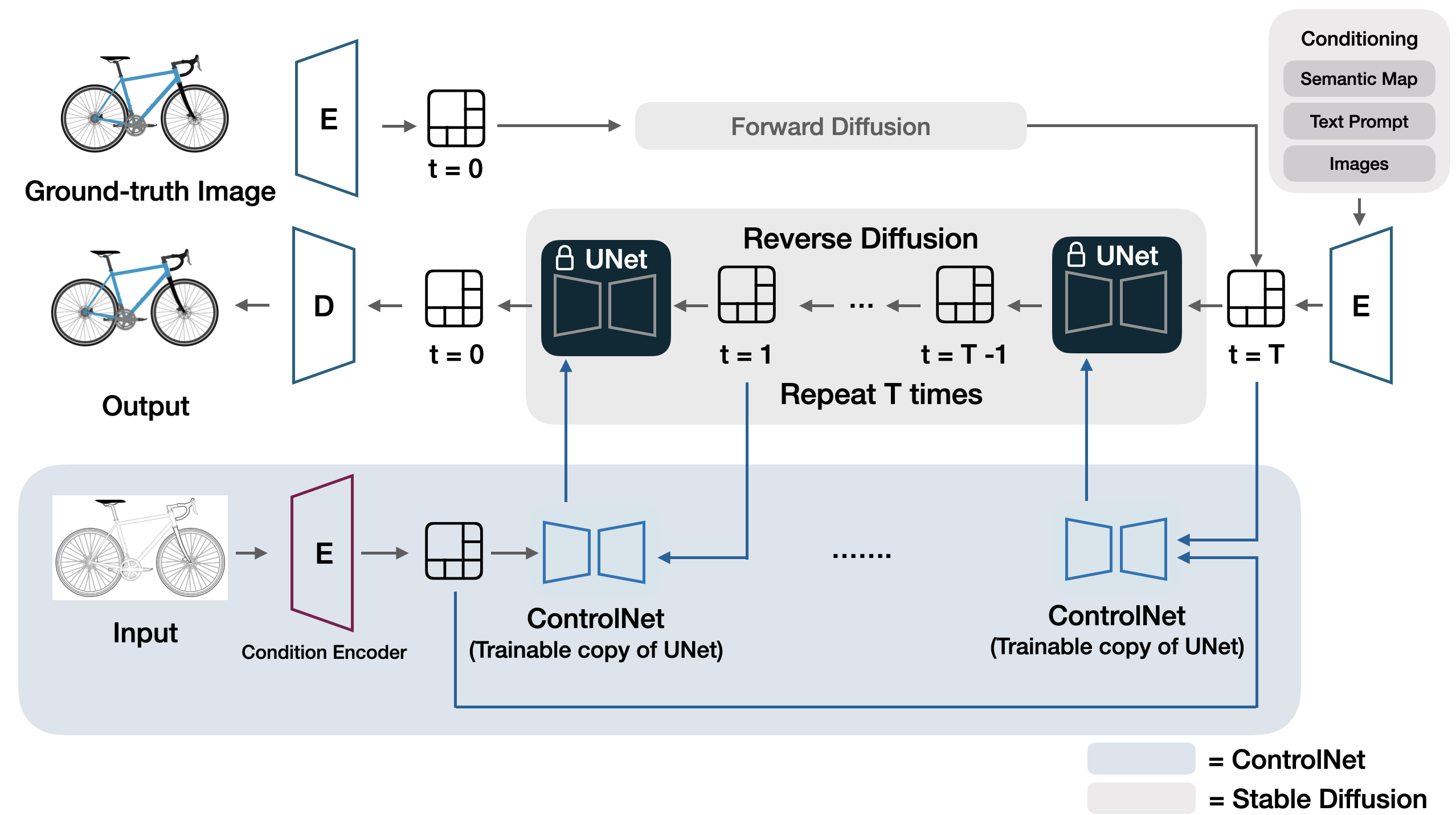}
    \caption{Overview of the ControlNet architecture.}
    \label{fig:controlnet}
\end{figure*}
\label{sec:related_work}

\label{sec:related_work}
The following sections delve into the evolution of generative models, multimodal control for generative models, and the limitations of state-of-the-art methods of using generative models for engineering design.

\subsection{Generative Models}
Recent years have witnessed significant progress in generative models, which aim to learn the underlying distribution of data and generate new samples that resemble the training data. Among the most prominent generative models are Generative Adversarial Networks (GANs)~\cite{GAN}, Variational Autoencoders (VAEs)~\cite{vae}, and Diffusion Models~\cite{diffusion}.
GANs, introduced by Goodfellow et al.~\cite{GAN}, consist of a generator network that learns to generate realistic samples and a discriminator network that distinguishes between real and generated samples. To address limitations such as mode collapse of traditional GANs, researchers have proposed various extensions and improvements, such as StyleGAN~\cite{stylegan} and BigGAN~\cite{biggan}. These models have also been extended to handle other data modalities, such as 3D shapes~\cite{3dgan} and video~\cite{ganvideo}.

VAEs, proposed by Kingma and Welling~\cite{vae}, learn a probabilistic encoding of the input data into a latent space and decode samples from this space to generate new data points. VAEs have been widely used for unsupervised learning and have been extended to handle various data modalities, such as text~\cite{vaetext} and graphs~\cite{simonovsky2018graphvae}. Recent works have focused on improving the sample quality and diversity of VAEs, such as VQ-VAE-2~\cite{vqvae2}, have achieved competitive results compared to GANs.

Diffusion models, first introduced by Sohl-Dickstein et al. in~\cite{diffusion}, have recently emerged as a powerful class of generative models. These models learn to generate samples by reversing a gradual noising process, starting from random noise and iteratively denoising it to obtain a clean sample. Diffusion models have shown impressive results in image generation~\cite{dhariwal2021diffusion}, outperforming GANs and VAEs in terms of sample quality and diversity. Stable Diffusion~\cite{sd} has gained significant attention due to its ability to generate high-quality images from textual descriptions, enabling a wide range of creative applications. Recent advancements in diffusion models include improved architectures, such as Diffusion Autoencoders~\cite{diffautoencoder} and Masked Autoencoder Diffusion Models~\cite{wei2023diffusion}, which have achieved state-of-the-art performance in image generation tasks. Diffusion models have also been applied to 3D shape generation~\cite{nichol2022point}, video generation~\cite{videosurvey}, and even music~\cite{huang2023noise2music}.
Despite the impressive capabilities of diffusion models in generating diverse and high-quality outputs, a significant challenge remains: the need for more precise control over the generation process. For example, in product design, developers may need to tweak specific features like size or texture without altering the overall structure of an image.\cite{saadi2023generative}. This gap highlights the necessity for mechanisms that can direct these models to produce specific, customized outputs required in specialized applications.

\subsection{Multimodal Control and Leveraging Foundation Models}
One of the significant drawbacks of generative models is the difficulty of controlling over the generated content. Various approaches have been proposed to add control to diffusion models, enabling more guided generation and enabling the ability to leverage pre-trained models, instead of training from scratch. This is even more significant for engineering design, where the amount of labeled and cleaned dataset is in high deficiency. ControlNet~\cite{zhang2023adding}, illustrated in Figure~\ref{fig:controlnet}, introduces a framework for adding conditional controls to pretrained diffusion models, allowing users to modify specific attributes of the generated content. In particular, it creates two copies of stable diffusion layers. One is locked, and the other is trainable and is conditioned on another input modality such as images. The trainable copy of the network also contains zero convolution and the results of the two networks are combined for each layer. This allows computationally efficient training and robustness to overfitting as the weights of the original Stable Diffusion model are locked~\cite{zhang2023adding}. Further, this architecture enables the possibility of conditioning the generative process in inputs such as edge maps for architecture, human pose graph for specific human motion generation. Further, this architecture allows adding control modalities to pre-trained foundation models, reducing the need for large amounts of data. Works are investigating the performance of ControlNet with different conditions. For example, Ju et al. address the issue of the uncertain and inconsistent image generation from ControlNet conditioning human pose, and proposes the HumanSD framework with greater accuracy~\cite{ju2023humansd}. Other methods focus on controlling the self-attention mechanism in diffusion models~\cite{hoe2023interactdiffusion}, enabling fine-grained control over the generation process. Additionally, techniques like classifier-free guidance~\cite{ho2022classifier} and contrastive language-image pre-training~\cite{crowson2022vqgan} have been employed to enhance the controllability of diffusion models. These approaches leverage additional information, such as text embeddings or classifier gradients, to guide the generation process toward desired outputs. 
Despite the advancements in multimodal learning and control, current methods primarily focus on integrating text and image modalities for controlling generative models, engineering products often involve a wide range of modalities, including parametric data, geometric constraints, assembly instructions, and performance requirements, which are not easily represented through text or images alone. Further, most works focus on training a novel architecture from scratch, which is infeasible for many engineering design problems, where large amounts of labeled data are difficult to obtain.

\subsection{Multimodal Generative Model for Engineering Design}
\begin{figure*}[t]
    \centering
    \includegraphics[width=17cm]{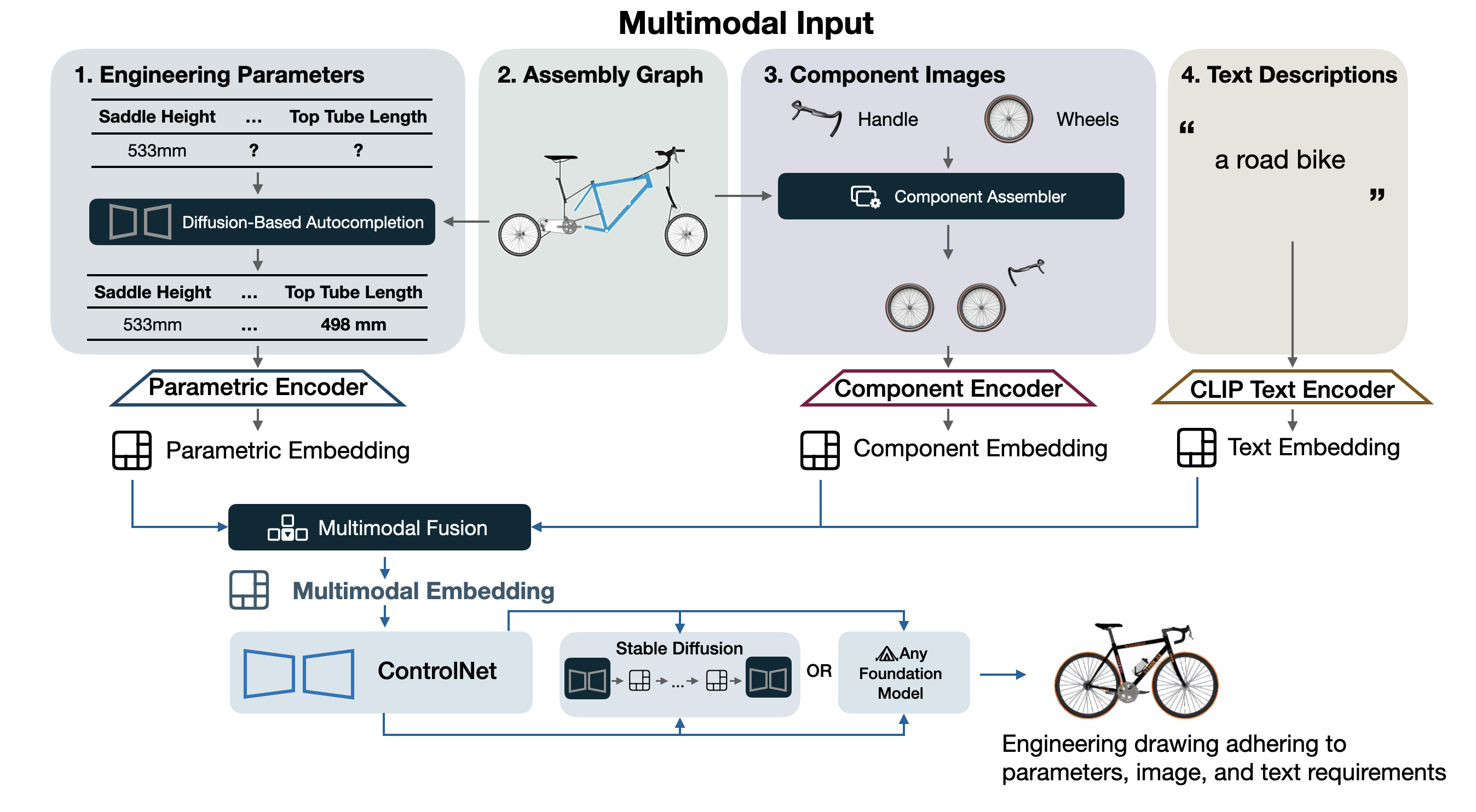}
    \caption
    {Overview of our model. We propose to add a parametric autocompletion and encoder conditioned on tabular data and assembly graphs, a component image encoder, a multimodal fusion module with attention layers to encode the parametric and textual information into a multimodal design parameter embedding.}
    \label{fig:overview}
\end{figure*}
Generative models and multimodal learning have shown significant potential in revolutionizing the field of engineering design by enabling the automatic creation, optimization, and evaluation of designs based on specified requirements and constraints. For example, Su et al. proposed a multimodal machine learning framework for predicting car ratings in ~\cite{10.1115/DETC2023-115076}. In recent years, diffusion models have emerged as a promising approach for engineering design synthesis and optimization. Prior research like ShipGen~\cite{bagazinski2023shipgen} and TopoDiff~\cite{maze2023diffusion}  demonstrate the effectiveness of diffusion models in generating optimized designs for various engineering domains, such as ship hull design and topology optimization. These models learn the gradual refinement process of design generation, enabling the creation of designs that meet specific performance targets and constraints. 
Generative models have also been applied to specific engineering design domains, such as materials design and microstructure optimization. For example, DiffMat~\cite{yuan2024diffmat} utilizes diffusion models to generate novel materials with desired properties, while work by Lee et al.~\cite{lee2024data} employs a diffusion-based approach to optimize the microstructure of materials for enhanced performance. Further, in~\cite{imputation}, Zhou et al. reimagined diffusion-based tabular imputation models as a way to enable parametric design autocomplete copilots. In~\cite{bagazinski2023shipgen} Bagazinski et al. utilized a conditional guided diffusion model for parametric ship hull design. In~\cite{edwards2024sketch2prototype}, Edwards proposed a pipeline that can transform hand-drawn sketches into 2D and 3D designs. Chong et al.~\cite{chong2024cad} explored CAD prompts for feasible and novel design generation. 
Despite the advancements in generative models for engineering design, there remain challenges and limitations. One major challenge is the incorporation of complex design modalities into the generative process. Engineering design often involves a multitude of information modalities, such as parametric data, assembly graphs, and component information. Further, the ability to leverage pre-trained foundation models instead of training from scratch is crucial for ensuring the success of using generative models in engineering design. 
For example, in automotive engineering, multimodal control would enable simultaneous adjustments to a car’s aerodynamics and aesthetics based on specific performance simulations and design criteria, ensuring that the final product meets both functional and visual standards. Further, there are very few high quality datasets that include both aerodynamics and aesthetics data. 
Thus, in this work, we aim to combine the power of diffusion models with multimodal control inputs to enable a versatile and effective generative model for engineering design. By incorporating parametric data, assembly graphs, component inspiration images, and textual descriptions, our proposed model can capture both design intent and also engineering constraints and generate optimized designs that meet specified requirements and constraints.

\section{Methodology}
\label{sec: method}

\subsection{Dataset}
We apply our model to the bike design problem. The BIKED Dataset is introduced by Regenwetter et al. in~\cite{biked}. The dataset contains the CAD files and parametric information of 4500 individually designed bicycles. We augment the dataset by randomly sampling the valid ranges for existing features to create an augmented version of the dataset, resulting in 12,506 samples. We further do a random training test split with a 90-10 ratio. 11255 samples are used for training and 1251 samples are used for testing and reporting performance. For each data sample, there is (1) a parametric vector containing 222 features that form a complete parametric representation for the full bike, (2) rendered images of the parametric features and the image resolution is 512x512, (3) a text description of "bike, white background."
\subsection{Training}
We train the model with the following hyperparameter settings: (1) learning rate: $1\times10^{-5}$, (2) batch size: 4, (3) number of epochs: 100. We train the model on a server with a 4090 GPU, an Intel i-9-13900K CPU, and 64Gb memory.

\subsection{Overall Pipeline}
Our proposed methodology develops a generative model for engineering design that leverages diffusion models and multimodal control inputs. The overall pipeline consists of five components, as illustrated in Figure~\ref{fig:overview}.

\begin{enumerate}
    \item Parametric Encoder with Autocompletion: We employ a diffusion-based imputation model inspired by~\cite{imputation} to handle incomplete parametric information. This generative model generates diverse and complete parametric designs from partial parametric inputs. Specifically, it generates a complete parametric design from the partial design. When the model is given a complete parametric design, we bypass the complete parametric design generation stage. With the complete parametric design, we use two fully connected layers to encode the parametric information.
    \item Component Image Assembly and Encoding: Component images are assembled according to their size and position defined in the assembly graphs. The assembled image is then encoded using a component encoder comprised of 8 convolutional layers to form the component embedding. 
    \item Text Encoding: Textual descriptions are encoded using a CLIP model~\cite{clip}, similar to Stable Diffusion~\cite{sd}, to capture semantic information and design intent.
    \item Multimodal Fusion: Embeddings from the parametric encoder, component encoder, and CLIP model are each of dimension 4096. The parametric embedding and the component embedding are concatenated and then we use a fully connected layer to project them into a vector of 4096, which is then added to the ClIP embedding from Stable Diffusion.
    \item Multimodal ControlNet: The multimodal embedding is used as a conditional input to a ControlNet-like module as shown in  Figure~\ref{fig:overview} to control the outputs of the foundation model layer by layer, enabling fine-grained control over the generated designs.
\end{enumerate}
\subsection{Parametric Encoder with Autocompletion}
The parametric autocompletion module is inspired by~\cite{imputation}, which introduces a generative imputation model for completing missing parametric data in engineering designs. It combines graph attention networks (GATs) and tabular diffusion models to capture and impute complex parametric interdependencies from an assembly graph.
The parametric autocompletion module serves as the first stage of the pipeline. It takes incomplete parametric designs as input and generates diverse and complete parametric designs. When the input is a complete parametric design, we bypass the parametric autocompletion module. The output is then passed through a parametric encoder comprised of two fully connected layers to obtain a compact parametric embedding, which serves as one of the inputs to the multimodal fusion stage.
The autocompletion module provides flexibility and adaptability, allowing the generative model to handle incomplete parametric information effectively and provide design recommendations and autocomplete functionality based on the available parametric data.

\subsubsection{Component Assembly and Encoding}
The second stage of our methodology incorporates component inspiration images and assembly graphs into the generative model through component assembly and component encoding.\\
\textbf{Component Assembly:} 
The component assembly step utilizes the structural information provided by the assembly graphs to assemble the component inspiration images into a coherent representation. Each node in the assembly graph represents a component, and the edges define their connections, relative positions, and relative sizes. We designed an assembly algorithm illustrated in Algorithm~\ref{alg:component_assembly} that retrieves the corresponding component inspiration images and positions and scales them according to the size and position attributes specified in the assembly graph. Though layering the correct-sized component images, the result is a composite image representing the assembled design.\\ 
In this process:
\begin{enumerate}

    \item Each component image $I_v$ is adjusted according to the size attribute $s_v$ specified in the assembly graph.
    \item The positioning $p_v$ determines the exact location of $I_v$ on the canvas.
    \item The resized and positioned component images are layered sequentially to create a composite image $I_{comp}$ that represents the assembled design.
\end{enumerate}
\begin{algorithm}[t]
\caption{Component Assembly Algorithm}
\label{alg:component_assembly}
\begin{algorithmic}[1]
\Require Assembly graph $G = (V, E)$, component images $\{I_v : v \in V\}$, size and position attributes $\{(s_v, p_v) : v \in V\}$
\Ensure Assembled composite image $I_{comp}$

\State Initialize a blank canvas $I_{comp}$
\For{each node $v \in V$}
    \State Resize component image $I_v$ to the specified size $s_v$
    \State Place resized $I_v$ at the position $p_v$ on the canvas
    \State Overlay $I_v$ on $I_{comp}$
\EndFor
\State \textbf{return} $I_{comp}$
\end{algorithmic}
\end{algorithm}
\textbf{Component Encoding:} The assembled composite image is then encoded into a meaningful representation using a component encoder consisting of convolutional and linear layers. This architecture is designed following the structure outlined in~\cite{zhang2023adding}, which has been proven effective for component conditioning tasks. Specifically, the component encoder comprises 8 convolutional layers with the following configuration: 2 layers with a dimension of 16 and filter size 3, 2 layers with a dimension of 32 and filter size 3, 2 layers with a dimension of 96 and filter size 3, 1 layer with a dimension of 256 and filter size 3, and 1 final layer with a dimension of 319. This design mirrors the configuration used in the original ControlNet paper, ensuring compatibility and effective feature extraction.

The convolutional layers extract relevant features and patterns from the image, capturing the spatial and structural information of the assembled design. This architecture was selected due to its proven success in handling component-specific encoding. The use of this tailored encoder, as opposed to pre-trained image encoders like CLIP, was motivated by the need for task-specific feature extraction that captures detailed spatial relationships critical for engineering design applications.

\subsection{Text Encoder}
To encode the textual descriptions of the bikes, we retain the original ControlNet's CLIP embedding mechanism~\cite{clip} to allow the next step of multimodal embedding fusion, we project the embedding vector after CLIP into a subspace of dimension of 4096.
\subsection{Multimodal Fusion}
The Multimodal Fusion process is critical for synthesizing diverse data modalities—engineering parametric data, component assembly, and textual descriptions—into a unified design representation. After obtaining the embedding vectors for each of these modalities, which are each of dimension 4096, the parametric embedding and the component embedding are concatenated and projected into a 4096-dimensional vector using a fully connected layer. This projected vector is then added to the CLIP embedding obtained from the pretrained Stable Diffusion model to create an integrated multimodal representation.

The parametric imputation model is pretrained, as detailed in~\cite{imputation}, and the CLIP model is also pretrained. The rest of the network, including the fusion layers and the downstream generative process, is trained end-to-end. This training approach ensures that the multimodal embeddings are aligned and optimized for the overall generative task.

Validation of the multimodal fusion process has been demonstrated in prior work~\cite{imputation}\cite{clip}. Specifically, downstream tasks involving controlled design generation and multimodal conditioning have shown that using this fusion process leads to coherent, contextually accurate design outputs that adhere to input specifications. We rely on these prior validations to confirm the effectiveness of the approach in synthesizing a unified design representation.

\subsection{ControlNet Module}
After we obtain the multimodal conditional embedding, we feed the vector to a ControlNet-like module that acts as a modifier over foundation models, which is Stable Diffusion in our case. Here, we follow the same model architecture as in~\cite{hu2023videocontrolnet}.

\section{Evaluations}

\begin{table*}[t]
\centering
\footnotesize 
\renewcommand{\arraystretch}{1.3} 
\setlength{\tabcolsep}{6pt} 
\caption{Test accuracies of surrogate models for parametric regression and classification based on image design. All surrogate models are ResNet-19 architectures trained on the full training set. We use these models to evaluate the constraint adherence in generated images from T2I foundation models.}
\label{table:surrogate}
\begin{tabular}{lccccccc}
\toprule
\textbf{Parameter} & \textbf{Saddle Height} & \textbf{Seat Tube Length} & \textbf{Stem Angle} & \textbf{Top Tube Length} & \textbf{Head Tube Angle} & \textbf{\shortstack{Number of \\ Water Bottles}} & \textbf{\shortstack{Handle \\ Bar Style}} \\
\midrule
$R^2$ Score & 0.97 & 0.98 & 0.95 & 0.98 & 0.96 & -- & -- \\
Accuracy & -- & -- & -- & -- & -- & 0.94 & 0.93 \\
\bottomrule
\end{tabular}
\end{table*}

\label{eval}
Our model extends the capabilities of generative design by enabling applications that existing state-of-the-art models cannot achieve effectively. Specifically, we demonstrate that models such as Stable Diffusion~\cite{sd} struggle to handle engineering-specific tasks where strict adherence to input parameters and complex component relationships are required. These limitations are evident when generating designs based solely on detailed parametric constraints or integrating component assembly information into the output.

To illustrate the strengths of our proposed pipeline, we conduct thorough quantitative and qualitative evaluations on the following tasks:
\begin{enumerate}
    \item \textbf{Accurate design generation conditional on parametric data only}: We assess how well the model generates complete designs that strictly adhere to input parametric specifications, validating this through metrics such as R\textsuperscript{2} scores and mean squared error comparisons against a surrogate model trained for parametric prediction.
    \item \textbf{Precise control over generated design using component images}: We evaluate the model's ability to generate designs that conform closely to provided component images, using metrics such as Intersection over Component (IoC) scores to measure fidelity to the input visual constraints.
    \item \textbf{Multimodal control involving both parametric data and component images}: We analyze how the model integrates and balances information from both parametric inputs and component images to produce cohesive outputs that respect both data sources. This is assessed through controlled experiments where different combinations of inputs are provided, and results are validated through surrogate model evaluations and visual inspection.
\end{enumerate}

In addition to standard image reconstruction quality metrics such as Peak Signal-to-Noise Ratio (PSNR) and Structural Similarity Index (SSIM), which are commonly used in prior works such as~\cite{sd}, we introduce a comprehensive evaluation framework that also includes a Diversity Score. 

PSNR is used to measure the fidelity of the generated bike designs compared to the expected design derived from parametric and component inputs. A higher PSNR value indicates that the generated image closely aligns with the specified design features, such as frame dimensions, wheel size, and component placements, ensuring minimal deviation from the input specifications. This metric is essential for verifying that the model can accurately translate input data into high-quality visual outputs. The PSNR is defined as:
\begin{equation}
\text{PSNR} = 10 \cdot \log_{10} \left(\frac{MAX_I^2}{\text{MSE}}\right),
\end{equation}
where $MAX_I$ represents the maximum possible pixel value of the image, and $\text{MSE}$ is the mean squared error between the original and generated images.

SSIM evaluates the perceptual quality of the generated designs, focusing on the structural similarity between the output and the expected design based on the parametric and component input. This metric considers changes in structural information, luminance, and contrast, making it more aligned with human visual perception. In the context of bike design generation, SSIM ensures that the model preserves the relationships between key components such as the frame, wheels, handlebars, and seat, leading to realistic and coherent outputs. The SSIM is defined as:
\begin{equation}
\text{SSIM}(x, y) = \frac{(2\mu_x \mu_y + C_1)(2\sigma_{xy} + C_2)}{(\mu_x^2 + \mu_y^2 + C_1)(\sigma_x^2 + \sigma_y^2 + C_2)},
\end{equation}
where $\mu_x$ and $\mu_y$ are the mean values of $x$ and $y$, $\sigma_x^2$ and $\sigma_y^2$ are the variances, $\sigma_{xy}$ is the covariance, and $C_1$, $C_2$ are constants to stabilize the division.

The Diversity Score measures the variability among generated outputs. In bike design generation, this metric is crucial for evaluating the model’s ability to produce a range of unique designs rather than repetitive or overly similar outputs. This diversity is particularly important in applications such as using the model as a design recommendation system. The Diversity Score is computed as:
\begin{equation}
\text{Diversity Score} = \frac{2}{n(n-1)} \sum_{i=1}^{n-1} \sum_{j=i+1}^{n} d(f(x_i), f(x_j)),
\end{equation}
where $n$ is the number of samples, $f(x_i)$ and $f(x_j)$ are feature representations of samples $x_i$ and $x_j$, and $d(\cdot, \cdot)$ denotes the distance measure between the feature representations.

By employing PSNR, SSIM, and the Diversity Score, we provide a comprehensive evaluation framework that ensures the generated bike designs not only match the input specifications accurately but also maintain high perceptual quality and design diversity.

\subsection{Surrogate Model for Parametric Generation Accuracy Analysis}

Measuring how well the model is responding to parametric conditions is a crucial step in the evaluation of the model. To provide a more rigorous analysis beyond visual inspection, we train a ResNet-18 model for each hand-selected feature that are of prominent visual difference, including both continuous and categorical features. Specifically, we use the feature-specific ResNet models as surrogate models to evaluate how accurately the generative pipeline adheres to these parametric conditions.

The features evaluated include:
\begin{enumerate}

    \item \textbf{Continuous features}: Saddle Height, Seat Tube Length, Stem Angle, Top Tube Length, and Head Tube Angle.
    \item \textbf{Categorical features}: Number of Water Bottles and Handle Bar Style.
\end{enumerate}

For each feature, a ResNet-18 model is trained to predict its value based on input images. The $R^2$ values are reported for regression tasks involving continuous features, while accuracy scores are reported for classification tasks involving categorical features, as shown in Table~\ref{table:surrogate}. To evaluate the pipeline's performance, we run inference on the generated images and compare the predicted parametric values from the surrogate models with the original input parameters. This quantitative analysis demonstrates the pipeline's ability to maintain accurate parametric conditioning across different feature types.

\subsection{Component Conditioning Analysis}
Further, evaluating how well the model is conforming to component images is also very important. To evaluate the pipeline's ability to follow the conditions on component images and assembly graphs, we compare the generated bikes against the assembled components and also evaluate the overall generation quality by manual inspection and quantitative metric evaluation. Specifically, we are evaluating how closely the model follows the assembled component images and if the model can generate valid and high-quality bikes using these conditions. We further demonstrate the model's ability by giving it out-of-distribution component images from the Internet and evaluating its performance.
\subsection{Multimodal Design Generation Analysis}
\begin{figure*}[t]
\centering
\includegraphics[width=\textwidth]{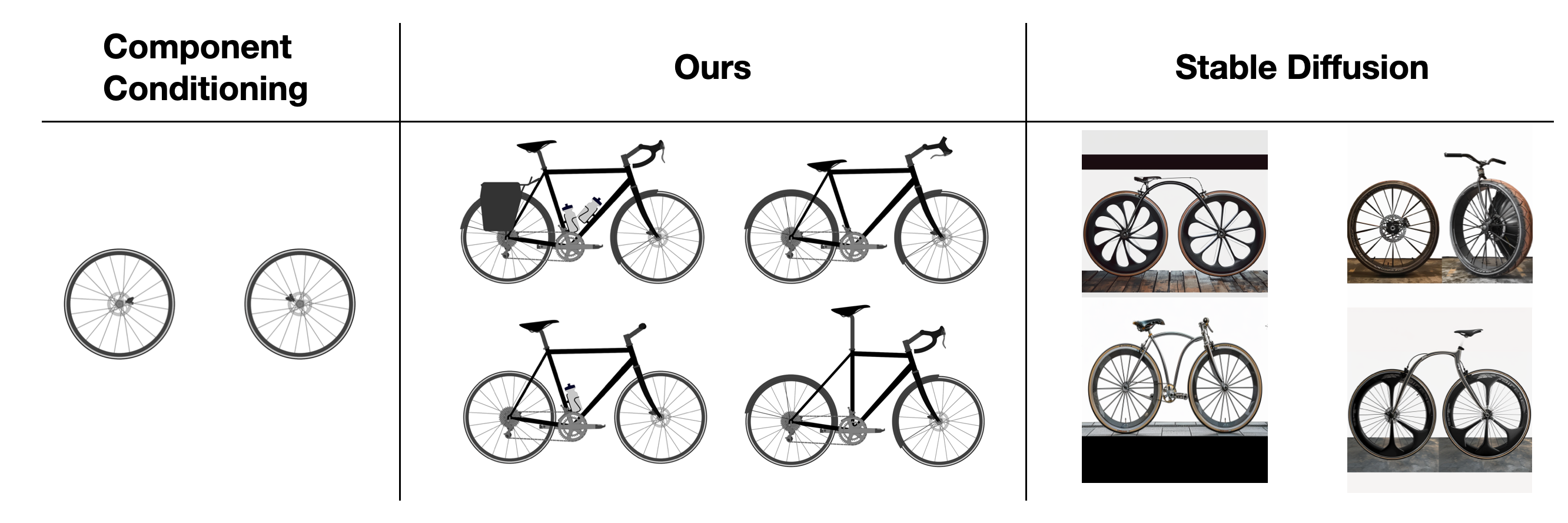}
\caption{Comparisons of generated results from Stable Diffusion vs. our model when conditioned on component information. We show 4 samples for both models. The inputs for both models are text and component images. The results indicate that Stable Diffusion struggles to generate feasible bikes under this condition, while our model successfully generates valid bikes with greater diversity.}
\label{fig:stable_diffusion_fail}
\end{figure*}

To evaluate the model's capability on multimodal conditioning on both parametric and component modalities, we first evaluate the model's ability to follow each of the modalities. We further evaluate the model in two adversarial cases: (a) when different modalities provide complementary information and we expect the model to conform to information contained in both modalities and (b) when the different modalities contain conflicting information and we expect one modality to overtake another modality. We evaluate the model using surrogate models and metrics defined above to evaluate the model over these situations with the metrics mentioned above to show the model's robustness under different modality overlapping cases.

\section{Experiment Results and Discussion}
\subsection{Shortcomings of Existing Models}
We demonstrate that our model can generate designs accurately controlled by various modalities, a capability that state-of-the-art models struggle to achieve. For example, Stable Diffusion, which represents one of the most popular T2I models, fails when tasked with multimodal generation based on parametric and component data. In our experiments, we embedded bike design parameters, such as size, structure, and specific features, directly into text prompts for Stable Diffusion. 

For example, we structure the text prompts for Stable Diffusion as follows:
\begin{itemize}
    \item For parametric-based generation: \texttt{"A bike design with a saddle height of 750mm, a seat tube length of 550mm, a stem angle of 15 degrees, a top tube length of 600mm, and a head tube angle of 73 degrees."}
    \item For component-based generation: \texttt{"A bike design featuring a frame that looks like the attached image."}
\end{itemize}

Figures~\ref{fig:stable_diffusion_fail1} and ~\ref{fig:stable_diffusion_fail} illustrate the limitations of Stable Diffusion in these contexts. Our findings indicate that while Stable Diffusion excels at general text-to-image generation, it struggles to generate valid bike designs that precisely adhere to specific parametric or component-based constraints. By contrast, as shown in the figures, our model successfully generates bike designs that align with the specified parametric and component-based conditions, demonstrating the effectiveness of our multimodal approach.

\subsection{Complete Parameter Set Generation}
\begin{figure}[t]
\centering
\includegraphics[width=8cm]{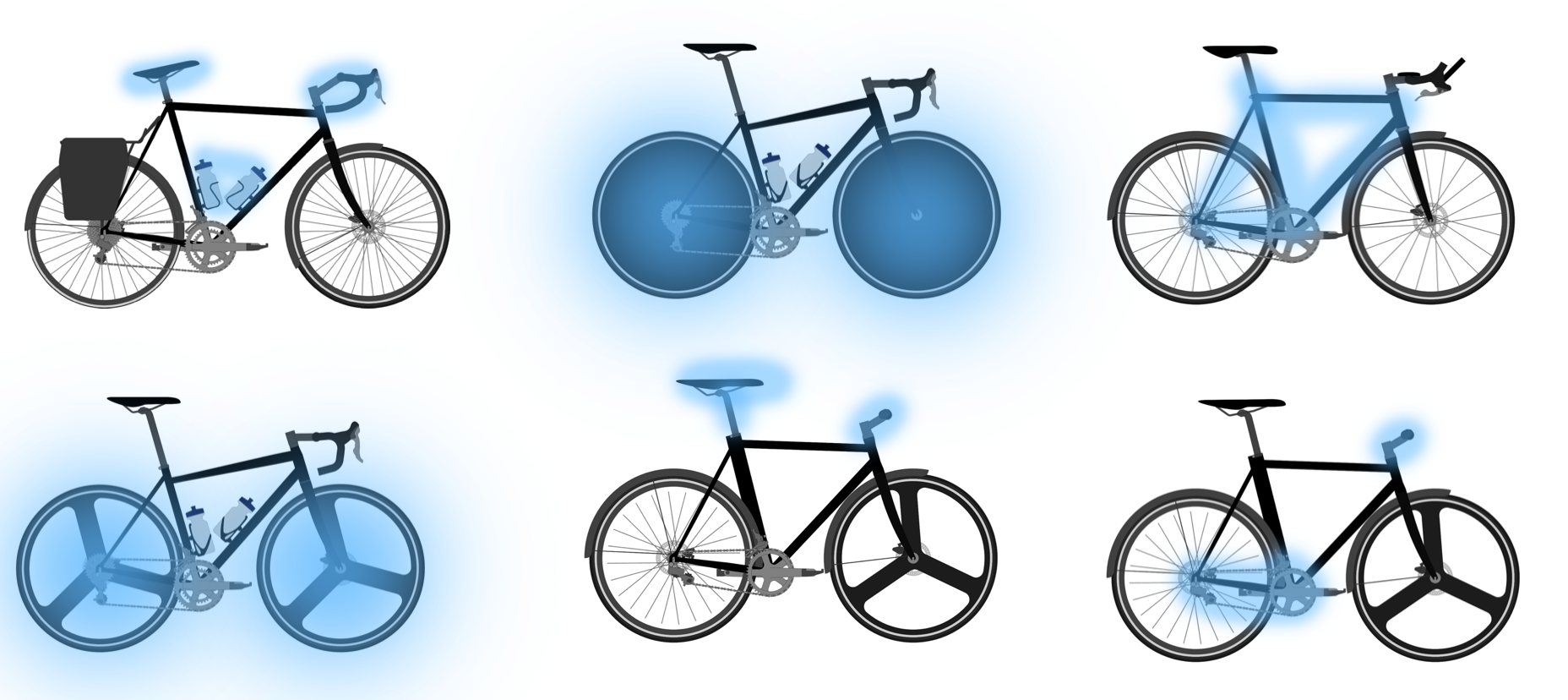}
\caption{Bike Renderings from Generated Parameters of the Imputation Model. Generated features are marked with a light blue tint. The input is testing parametric information with 10\% of features masked. We then use the imputation model to generate the missing features. We show these hand-picked samples to show that the imputation model generates valid and diverse bike designs.}\label{fig:imputation model}
\end{figure}

We first conduct a comprehensive analysis on the accuracies of parametric controlled generation. We show it is effective and enables the capability of design generation conditioned on an arbitrary subset of the parameters. First, we show some of the generated bikes from generated features of the imputation model in Figure 
\ref{fig:imputation model}.

\begin{table*}[t]
\label{Table: imputation ablation}
\centering
\renewcommand{\arraystretch}{1.2} 
\setlength{\tabcolsep}{12pt} 
\caption{Ablation study evaluating the impact of using the imputation model from~\cite{imputation} for partial parametric conditioning. The study compares two methods for handling missing features: (a) employing the imputation model, and (b) providing a feature vector with missing values set to '0', accompanied by a masking tensor that indicates which values are missing. The results show that the imputation model (approach a) consistently achieves superior performance across SSIM, PSNR, and Diversity Score metrics. The evaluation is based on a sample size of 1000. Metrics marked with $\uparrow$ indicate that higher values are better, and bold text highlights the best performance for each metric.}
\begin{tabular}{lccc}
\toprule
\textbf{Model} & \textbf{SSIM} $\uparrow$ & \textbf{PSNR} $\uparrow$ & \textbf{Diversity Score} $\uparrow$ \\
\midrule
With Imputation Model & \textbf{0.83} & \textbf{27.1} & \textbf{0.16} \\
Without Imputation & 0.78 & 25.8 & 0.12 \\
\bottomrule
\end{tabular}
\label{tbl:ablation}
\end{table*}

\subsubsection{Ablation Study of Imputation Model}
To demonstrate the effectiveness of using an imputation model for parametric data completion, we conduct an ablation study, as shown in Table~\ref{tbl:ablation}. The study compares performance metrics when employing the imputation model described in~\cite{imputation} against an alternative approach that feeds the downstream pipeline with: (a) a mask vector indicating missing features, and (b) a feature vector where missing features are set to zero. The results indicate that incorporating the imputation model significantly improves the model's performance, enhancing both accuracy and diversity in the generated outputs.

\begin{table*}[t]
\centering
\small
\renewcommand{\arraystretch}{1.3} 
\setlength{\tabcolsep}{10pt} 
\caption{Performance of our model's ability to condition on parametric information as measured by $R^2$ using surrogate models. The table shows mean values and standard deviations. We compare the model's performance under two different modes of providing parametric information: (a) employing the imputation model, and (b) providing a feature vector with
missing values set to ’0’, accompanied by a masking tensor that indicates which values are missing.}
\label{table:component_acc}
\begin{tabular}{lccccc}
\toprule
\textbf{Feature} & \textbf{Saddle Height} & \textbf{Seat Tube Length} & \textbf{Stem Angle} & \textbf{Top Tube Length} & \textbf{Head Tube Angle} \\
\midrule
\textbf{Ours-Imputation} & 0.87 (0.15) & 0.89 (0.16) & 0.89 (0.14) & 0.91 (0.11) & 0.88 (0.19) \\
\textbf{Ours-Masking} & 0.85 (0.16) & 0.79 (0.11) & 0.82 (0.19) & 0.90 (0.21) & 0.86 (0.20) \\
\bottomrule
\end{tabular}
\end{table*}

\begin{table*}[t]
\centering
\small
\renewcommand{\arraystretch}{1.3} 
\setlength{\tabcolsep}{6pt} 
\caption{Performance evaluation of our model's ability to condition on parametric information of selected features. We analyze the accuracy of generated feature values compared to input parametric data using surrogate models. The experiment involves randomly selecting 10 bikes from the test dataset, masking out the specific feature being studied, generating 100 complete bike design samples, and measuring the corresponding feature values from the output images. Median values are close to the target parameter values, showing that the model respects the constraints when it is conditioned on parametric information.}
\label{table:parametric_specific_feature}
\begin{tabular}{lcccc}
\toprule
\textbf{Feature} & \textbf{Water Bottle = 2} & \textbf{Seat Tube Length = 678mm} & \textbf{Rack = 1} & \textbf{Front Render Angle = 60} \\
\midrule
\textbf{Median} & 2.0 & 675.0 & 1.0 & 62.0 \\
\textbf{Average} & 1.95 & 673.56 & 0.98 & 62.34 \\
\textbf{Standard Deviation (SD)} & 0.1 & 15.7 & 0.01 & 2.67 \\
\textbf{\% Error} & N/A & 0.18\% & N/A & 0.07\% \\
\textbf{Accuracy} & 0.85 & N/A & 0.98 & N/A \\
\bottomrule
\end{tabular}
\end{table*}

\subsection{Parametric Generation}
\subsubsection{Using surrogate models to Evaluate Parametric Conditioning}
Further, to closely evaluate the model's ability to accurately follow parametric conditions, we use the surrogate model to predict the parameters of the generated designs against the original parameter set for 10 randomly selected features that include both categorical and continuous features. Specifically, we use our model in parametric conditioning only mode: the inputs are complete parametric vectors and we use the model to generate the final design based on the parametric information. We then use the surrogate models to predict the parametric information based on the generated designs. Since we give the model complete sets of parametric designs as inputs, we expect the model to produce designs that accurately reflect the values for the features in the parametric space. For all tests, we show the values of $R^2$ for continuous features and accuracy scores for categorical features in Table~\ref{table:component_acc}. We compare the model's performance under two different modes of providing parametric information: (a) employing the imputation model, and (b) providing a feature vector with
missing values set to ’0’, accompanied by a masking tensor that indicates which values are missing. Further, we conduct a comprehensive evaluation study on parametric conditioning on specific features to examine the model's behavior more closely. The results are shown in Table~\ref{table:parametric_specific_feature} and indicate the model is able to generate parametrically accurate designs based on both numerical and categorical features.


\begin{figure}[t]
\centering
\includegraphics[width=8cm]{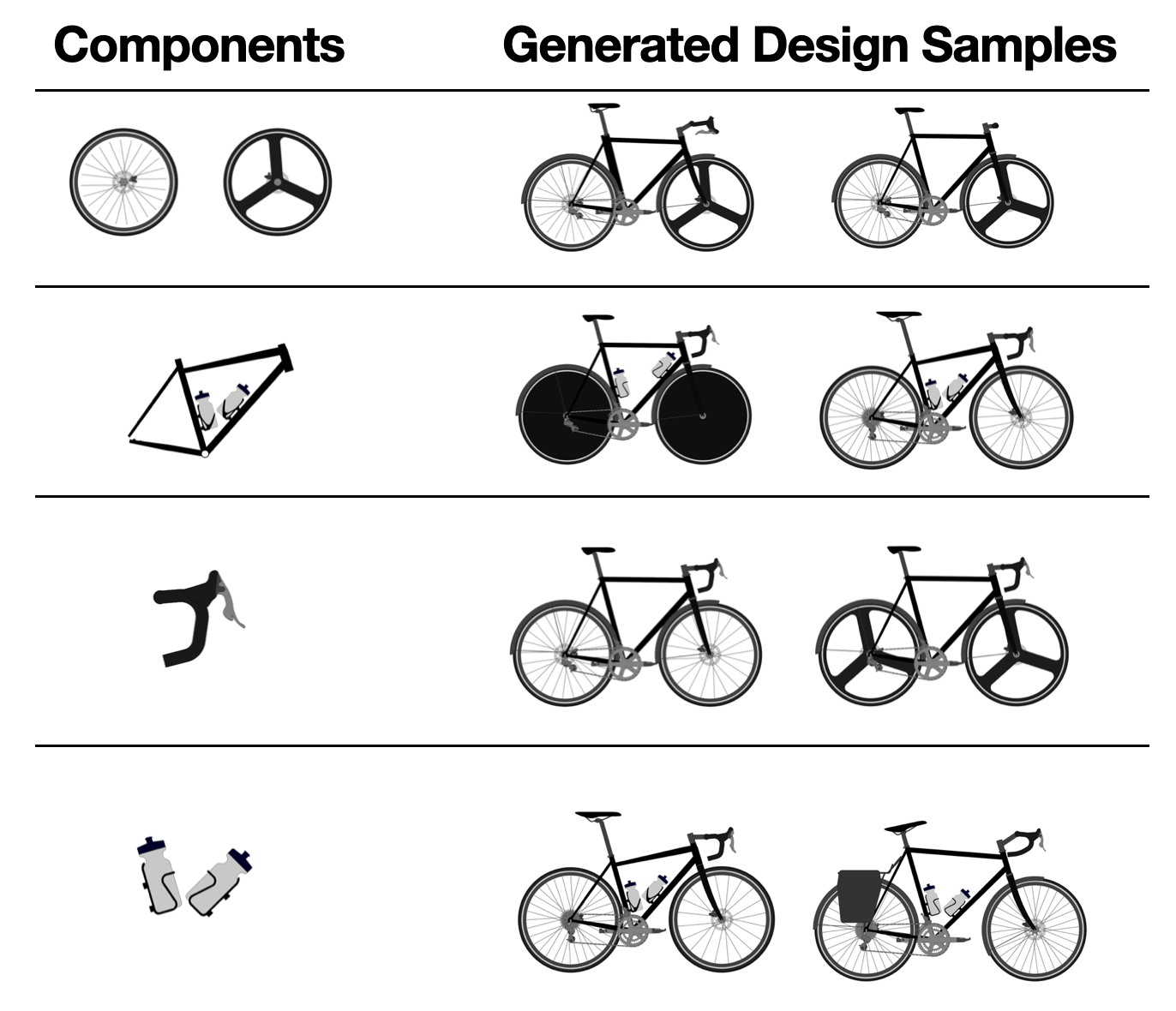}
\caption{Sample Results when using our model in the component conditioning mode. We observe that generated design samples also contain components similar to the input component image.}\label{fig:component}
\end{figure}

\subsection{Component Conditioning}
\begin{figure}[t]
\centering
\includegraphics[width=8.5cm]{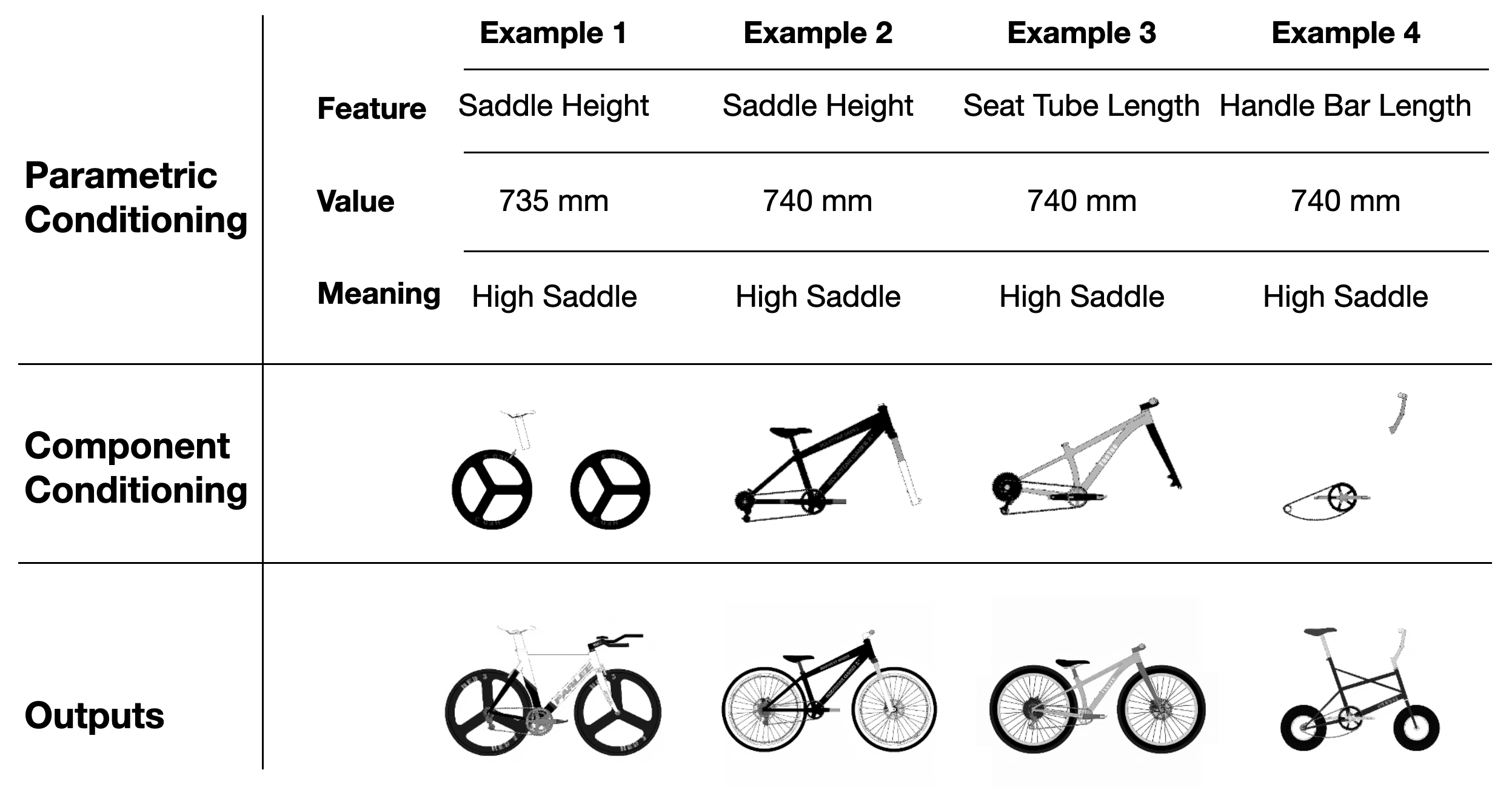}
\caption{Model's Output When Given Conflicting Information between Different Modalities.  We observe that generated design samples contain components more similar to the input image of components, ignoring the parametric conditioning.}\label{fig: multimodalconflicting}
\end{figure}

\subsubsection{Evaluations on Component Conditioning}
Further, we are interested in thoroughly evaluating the model's ability to condition on component images. We provide the model with component images of a predefined subset of seven components. The number of components present in the component conditioning stage can be anything between 1 and 7. We show a random set of generated results in Figure~\ref{fig:component}.
Further, generated designs were subsequently evaluated for their fidelity to these component inputs. To quantify the similarity between the input component and the corresponding section of the generated image, we established a ratio: Intersection over Component (IoC), defined as:
\begin{equation}
    IOC=\frac{Area_{intersection}}{Area_{component}},
\end{equation}. Beyond mere fidelity, the diversity across the generated set was computed to understand the model's variability in output for a given input. We show the results in Table~\ref{table:single_component}.

\begin{table*}[t]
\centering
\small
\renewcommand{\arraystretch}{1.3} 
\setlength{\tabcolsep}{8pt} 
\caption{Quantitative Analysis of Single Component Image Conditioning. For each feature, 100 samples were tested. Metrics include Intersection over Condition (IoC), Peak Signal-to-Noise Ratio (PSNR), and Structural Similarity Index Measure (SSIM). Higher values indicate better performance. High IoC values for all parameters indicate that the samples generated by our model have component images similar to the input component image.}
\label{table:single_component}
\begin{tabular}{lccccc}
\toprule
\textbf{Metric} & \textbf{Crank} & \textbf{Bottle} & \textbf{Saddle} & \textbf{Handle Bar} & \textbf{Frame} \\
\midrule
\textbf{IoC} $\uparrow$ & 0.78 & 0.85 & 0.82 & 0.84 & 0.85 \\
\textbf{PSNR} $\uparrow$ & 0.46 & 0.48 & 0.52 & 0.47 & 0.48 \\
\textbf{SSIM} $\uparrow$ & 11.9 & 12.2 & 12.5 & 11.3 & 11.5 \\
\bottomrule
\end{tabular}
\end{table*}


\begin{table*}[t]
\centering
\small
\renewcommand{\arraystretch}{1.3} 
\setlength{\tabcolsep}{8pt} 
\caption{Non-overlapping multimodal conditioning evaluation. For each of the five cases listed, we randomly selected 10 bikes from the dataset and removed the corresponding components from the parametric condition while providing a conditioning image for this component. Our model generated 100 designs conditioned on both image and parametric modalities. High $R^2$ and IoC values demonstrate successful adherence to conditions in both modalities.}
\label{table:multimodal_non_overlapping}
\begin{tabular}{lccccc}
\toprule
\textbf{Metric} & \textbf{Crank} & \textbf{Bottle} & \textbf{Saddle} & \textbf{Handle Bar} & \textbf{Frame} \\
\midrule
\textbf{IoC} $\uparrow$ & 0.78 & 0.85 & 0.82 & 0.84 & 0.85 \\
\textbf{$R^2$} $\uparrow$ & 0.85 & 0.86 & 0.88 & 0.90 & 0.87 \\
\textbf{PSNR} $\uparrow$ & 0.46 & 0.48 & 0.52 & 0.47 & 0.48 \\
\textbf{SSIM} $\uparrow$ & 11.9 & 12.2 & 12.5 & 11.3 & 11.5 \\
\bottomrule
\end{tabular}
\end{table*}

\begin{table*}[t]
\centering
\small
\renewcommand{\arraystretch}{1.3} 
\setlength{\tabcolsep}{8pt} 
\caption{Conflicting multimodal conditioning evaluation. For each of the three cases, we randomly selected 10 bikes from the dataset and added component image conditioning that conflicted with the corresponding features in the parametric condition. Our model generated 100 samples, evaluated using the surrogate model to assess the respective features in the generated designs. The results indicate that the model typically resolves conflicts by leaning towards the component features more. Values in parentheses represent standard deviation.}
\label{table:multimodal_conflicting}
\begin{tabular}{lcccc}
\toprule
\textbf{Features} & \textbf{Saddle Height (mm)} & \textbf{Seat Tube Length (mm)} & \textbf{Handle Bar Length (mm)} \\
\midrule
\textbf{Parametric Space} & 800 & 505 & 20 \\
\textbf{Component Space} & 550 & 540 & 100 \\
\textbf{Generated Designs} & 560 (15.6) & 546 (12.1) & 98mm (8.9) \\
\bottomrule
\end{tabular}
\end{table*}

\begin{figure*}[t]
\centering
\includegraphics[width=\textwidth]{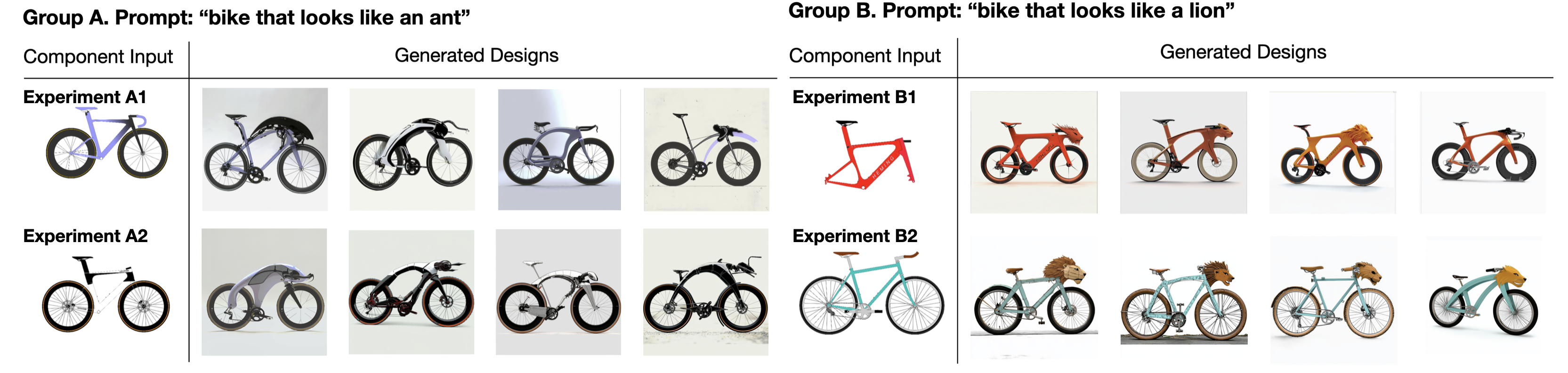}
\caption{Creative component conditioning experiments demonstrating the model's ability to retain creativity while incorporating multimodal information. The text prompts are shown in the figure, with the leftmost column displaying the component image constraints. The results illustrate how the model effectively integrates these constraints to generate diverse and creative outputs.}
\label{fig:creative_prompting}
\end{figure*}

\subsection{Multimodal Design Generation}
Furthermore, investigating how well the models behave under multimodal conditioning inputs is also very important. For each of the following cases, we evaluate the model's performance when given multimodal information including parameters, component images, and assembly graphs. In each of the cases, we evaluate the model's performance by following each of the modalities.

\subsubsection{Non-Overlapping Cases}
In complex engineering tasks, different modalities often provide unique and complementary information. To test the model's ability to integrate these distinct streams effectively, this experiment uses parameters and component images as non-overlapping inputs. Each modality independently contributes specific information that, when combined, forms the complete set of design specifications. This setup assesses the model's capability to synthesize information from disparate sources to produce coherent and complete designs. 
To test this, we mask the parameters to ensure no redundancy with component images. However, when put together, the parameters and the component images represent the complete parameter set. We show the results in Table~\ref{table:multimodal_non_overlapping}. 

\subsubsection{Adversarial Case: Conflicting Information}
In real-world design scenarios, engineers and designers often face conflicting requirements or constraints that must be carefully balanced or prioritized. To simulate such challenges and evaluate the model’s resilience and decision-making capabilities, we intentionally introduce conflicting constraints into our experiment. For example, we provide the model with parametric information indicating a high saddle while providing a component image indicating a low saddle. This setup tests the model's ability to handle discrepancies and prioritize information from different modalities when they conflict.
The results, illustrated in Figure~\ref{fig: multimodalconflicting} and Table~\ref{table:multimodal_conflicting}, show that the model tends to prioritize component images. This is expected because although the model is trained on both parametric data and component images, the primary training loss is based on image reconstruction. This emphasis on image reconstruction encourages the model to prioritize the visual information from component images, which directly affects the quality and fidelity of the generated output. Component images inherently provide richer spatial and structural cues, making them more influential in guiding the model's decisions compared to the abstract, numerical nature of parametric data. This behavior ensures that the generated designs maintain high visual coherence, even when input modalities conflict.

\subsubsection{Creativity Assessments}

A key challenge when integrating parametric or image-based controls into generative models like Stable Diffusion is the potential loss of creative output, which is a defining strength of foundational models. To address this concern, it's crucial to evaluate whether the model retains its capability to generate innovative and unique designs that diverge from existing images in the BikeCAD dataset. This section assesses the model's ability to produce creative and contextually relevant designs while adhering to specific multimodal inputs.
We conducted three experiments using unique text prompts to guide the generation process, aiming to test the model's capacity to blend thematic creativity with technical design constraints. Specifically, the inputs are the component images and the text descriptions shown below and the outputs are the generated designs. We generate 30 samples for each case and hand-pick the four most prominent ones.
The results, as shown in the figure, demonstrate the model's capability to integrate thematic elements into the design of bicycles. The results are shown in Figure~\ref{fig:creative_prompting}.

A. Text Prompt: ``bike that looks like an ant''
The resulting designs showcase the model's proficiency in capturing the essence of an ant's form, including segmented structures and compact, robust frames. The model successfully generates bike designs that have ant elements.

B. Text Prompt: "bike that looks like a lion"
In this experiment, we explore the model's ability to generate bike designs that resemble a lion. The model successfully generates bike designs that have elements of a lion.

To quantitatively evaluate the model’s creativity and adherence to inputs, we used the CLIP model to measure the average distances of generated samples to the input component image, the text prompt, and the training dataset, as shown in Table~\ref{table:creative_clip}.

The low average distances to the input component image (0.13 to 0.16) indicate that the model maintains structural consistency, ensuring designs that remain true to the technical components. Distances to the text prompts (0.55 to 0.61) show that the model captures thematic elements, allowing it to reflect the “ant” and “lion” concepts while introducing creative variations. Finally, the distances from the training dataset (0.17 to 0.21) confirm that the generated designs are sufficiently distinct, highlighting the model's capability to produce novel outputs.

These results affirm that the model effectively combines technical fidelity with creative flexibility, generating unique, contextually relevant designs in response to thematic prompts.

\begin{table*}[t]
\centering
\small
\renewcommand{\arraystretch}{1.3} 
\setlength{\tabcolsep}{8pt} 
\caption{Quantitative Evaluation of the creative component conditioning experiments shown in Figure~\ref{fig:creative_prompting} using the CLIP model. We calculate the average distances of 100 generated samples to (1) the input component image, (2) the text prompt, and (3) bikes in the training dataset.}
\label{table:creative_clip}
\begin{tabular}{lcccc}
\toprule
\textbf{Avg CLIP Distance to} & \textbf{Input Component Image} & \textbf{Text Prompt} & \textbf{All Dataset Bikes} & \textbf{Random Bike in Dataset}\\
\midrule
\textbf{Experiment A1} & 0.13 & 0.61 & 0.21 & 0.20\\
\textbf{Experiment A2} & 0.14 & 0.59 & 0.19 & 0.16 \\
\textbf{Experiment B1} & 0.16 & 0.57 & 0.17 & 0.15 \\
\textbf{Experiment B2} & 0.15 & 0.55 & 0.20 & 0.18  \\
\bottomrule
\end{tabular}
\end{table*}
\section{Discussion}
Our model represents an advancement in engineering design by offering strong multimodal control over foundation models, particularly tailored for engineering applications. By integrating parametric inputs, assembly graphs, and component inspiration images, our model enables precise design generation with diversity. The discussion will focus on the implications of our model, its potential applications, and future directions for research and development.

\subsection{Implications and Applications to Engineering Design}
The introduction of our model has profound implications for engineering design across various industries. By providing designers with enhanced control over generated designs, our model facilitates the exploration of complex design spaces, enabling the creation of innovative and optimized solutions. Moreover, the ability to incorporate diverse data modalities such as parametric inputs, assembly graphs, and component images enhances the fidelity and accuracy of generated designs, ensuring alignment with specifications and constraints. This capability is particularly valuable in domains such as product design, architecture, and manufacturing, where precise control and customization are paramount.

Furthermore, our model opens up new opportunities for collaborative design processes, allowing multidisciplinary teams to leverage diverse expertise and insights. Designers can iteratively refine and explore design alternatives in a collaborative environment, incorporating feedback from stakeholders and domain experts. Additionally, the versatility of our model enables its integration into existing design workflows and platforms, enhancing productivity and streamlining the design process.

\subsection{Future Directions}
While our model represents a significant step forward in engineering design generation, there are several avenues for future research and development. One direction is the refinement and optimization of the model architecture to further enhance its performance and scalability. In particular, future work may involve exploring advanced multimodal fusion techniques, incorporating additional control modalities, such as engineering performance (dynamics, ergonomics, structural, aerodynamics, etc.), environmental factors (design for specific terrains), or 3D information conditioning through representations such as mesh and point cloud.

Another promising direction is the exploration of novel applications and use cases for our model across diverse domains. For example, our model could be applied to parametric CAD design problems in fields such as automotive engineering, aerospace, and biomedical device design. By adapting the model to specific domain requirements and constraints, researchers can unlock new opportunities for innovation and problem-solving.

Additionally, future research could focus on the development of tools and methodologies for evaluating and validating generated designs. This may involve the creation of benchmark datasets, design metrics, and evaluation protocols to assess the quality, diversity, and functionality of generated designs objectively. By establishing rigorous evaluation standards, researchers can ensure the reliability and robustness of generative models in real-world design scenarios.

Overall, the development and refinement of generative models for engineering design represent a promising frontier in AI-driven design research. By continuing to innovate and explore new possibilities, researchers can empower designers with powerful tools and methodologies for tackling complex design challenges and driving innovation in engineering.

\section{Conclusion}
We presented a novel generative model designed to exert multimodal control over text-to-image foundation models, specifically tailored for engineering design applications. By integrating parametric inputs, assembly graphs, and component inspiration images, our model offers precise design generation. Our work represents a significant advancement in AI-driven design tools, with implications for diverse industries and collaborative design processes. Moving forward, we envision continued research and development to refine and optimize our model, explore novel applications across different domains, and establish rigorous evaluation standards. Ultimately, our model has the potential to revolutionize engineering design by empowering designers with powerful tools and methodologies for innovation and problem-solving.

{\small
\bibliographystyle{ieee_fullname}

\bibliography{egbib}
}

\end{document}